\pdfoutput=1

\documentclass[11pt]{article}

\usepackage{acl}

\usepackage{times}
\usepackage{latexsym}

\usepackage[T1]{fontenc}

\usepackage[utf8]{inputenc}
\usepackage{times}
\usepackage{latexsym}
\usepackage{url}
\usepackage{enumitem}
\usepackage{amsmath,amssymb,amsfonts}
\usepackage{algorithmic}
\usepackage{caption}
\usepackage{subcaption}
\usepackage{graphicx}
\usepackage{textcomp}
\usepackage{url}
\usepackage{makecell}
\usepackage{multirow}
\usepackage{booktabs}
\usepackage{array}
\usepackage{microtype}
\usepackage{kotex}
\usepackage{tikz}
\usepackage{pgfplots}
\usepackage{xspace}
\usepackage{multirow}
\usepackage{booktabs}
\usepackage{hyperref}

\usepackage{hyperref}       
\usepackage{amsfonts}       
\usepackage{nicefrac}       
\usepackage{xcolor}         
\usepackage{tcolorbox}
\usepackage{fvextra}
\newlist{longenum}{enumerate}{1}
\setlist[longenum,1]{leftmargin=4mm, label=\roman*)}
\definecolor{lp}{HTML}{CBC3E3}
\definecolor{purple}{rgb}{0.56, 0.0, 1.0}
\usepackage{colortbl}

\newcommand{\thickmidrule}{%
    \noalign {\ifnum 0=`}\fi \hrule height 1pt
    \futurelet \reserved@a \@xmidrule
}

\usepackage{comment}
\usepackage{todonotes}
\usepackage{xcolor}

\usepackage{microtype}
\newcommand\blfootnote[1]{%
  \begingroup
  \renewcommand\thefootnote{}\footnote{#1}%
  \addtocounter{footnote}{-1}%
  \endgroup
}

\title{SAAS: Solving Ability Amplification Strategy for Enhanced Mathematical Reasoning in Large Language Models}

\author{Hyeonwoo Kim$^{1*}$, Gyoungjin Gim$^{1*}$, Yungi Kim$^{1*}$ \\ {\bf \large Jihoo Kim$^{1}$, ‪Byungju Kim$^{2}$‬, Wonseok Lee$^{2}$, } \\ {\bf \large Chanjun Park$^{1\dagger}$}\\
\\
  $^{1}$ Upstage AI, $^{2}$ Mathpresso Inc. \\
  \texttt{\{choco\_9966, gyoungjin.gim, eddie, jerry, chanjun.park\}@upstage.ai} \\
  \texttt{\{peyton.kim, jack.lee\}@mathpresso.com}}

\begin{document}
\maketitle
\begin{abstract}
\blfootnote{$^*$Equal Contribution $^\dagger$ Corresponding Author}
This study presents a novel learning approach designed to enhance both mathematical reasoning and problem-solving abilities of Large Language Models (LLMs). We focus on integrating the Chain-of-Thought (CoT) and the Program-of-Thought (PoT) learning, hypothesizing that {\it prioritizing the learning of mathematical reasoning ability is helpful for the amplification of problem-solving ability}. Thus, {\it the initial learning with CoT} is essential for solving challenging mathematical problems. To this end, we propose a sequential learning approach, named \textsc{\textbf{SAAS}} (Solving Ability Amplification Strategy), which strategically transitions from CoT learning to PoT learning. Our empirical study, involving an extensive performance comparison using several benchmarks, demonstrates that our \textsc{\textbf{SAAS}} achieves state-of-the-art (SOTA) performance. The results underscore the effectiveness of our sequential learning approach, marking a significant advancement in the field of mathematical reasoning in LLMs.
\end{abstract}

\section{Introduction}
The advent of Large Language Models (LLMs) has marked a significant breakthrough in various domains. However, despite their remarkable performance across these domains, a notable challenge persists in the realm of mathematical reasoning~\cite{survey_llm, survey_math, survey_mlp, qian2022limitations, zhou2022teaching, verify, drori2021neural, zhang2019gap}. The ability of LLMs to comprehend, interpret, and manipulate mathematical concepts is not yet on par with their linguistic capabilities.

The significance of mathematical reasoning in LLMs involves more than just crunching numbers. It also encompasses the ability to engage in logical thinking, problem-solving, and complex decision-making, which are essential for understanding and generating human-like responses in the different situations~\cite{survey_math, survey_mlp, survey_representing}. In other words, mathematical reasoning in LLMs is essential for a comprehensive understanding and manipulation of language in numerous scientific and practical applications. However, the current ability of LLMs in mathematical reasoning hinder their potential in the fields where numerical and logical comprehension are paramount such as coding. Thus, it's critical challenge to enhance the ability of LLMs in mathematical reasoning.

\begin{figure*}[t!]
    \centering
    \includegraphics[width=1.0\textwidth]{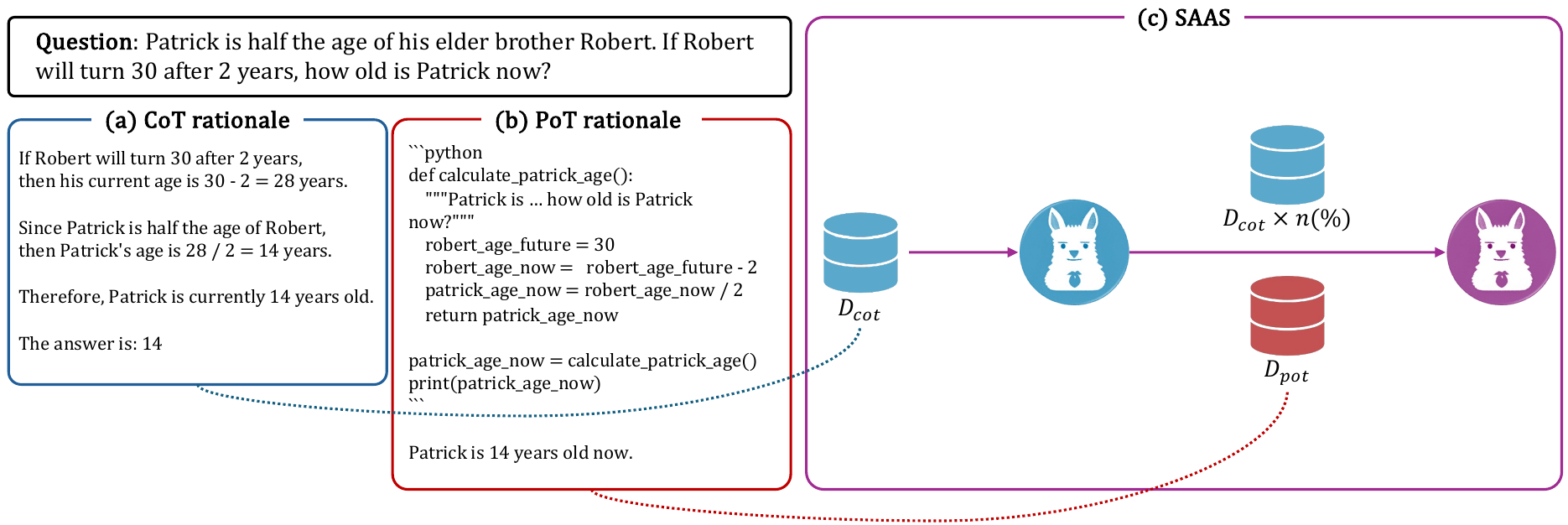}
    \caption{Overview of \textsc{\textbf{SAAS}} (Solving Ability Amplification Strategy) with two core strategies: i) sequential learning strategy; ii) cognitive retention strategy.}
    \label{fig:model}
\end{figure*}

In this study, we explore a learning approach for enhancing both mathematical reasoning ability and problem-solving ability in LLMs, focusing on learning with both the Chain-of-Thought (CoT)~\cite{cot} and the Program-of-Thought (PoT)~\cite{pot, pal}. The CoT rationale (Figure~\ref{fig:model}-(a)) consists of a series of intermediate reasoning steps. Although it enhances the reasoning ability of LLMs, it leads to arithmetic calculation errors when dealing with large numbers~\cite{pot}, resulting a low problem-solving ability. To address this issue, \citet{pot} proposed the PoT rationale (Figure~\ref{fig:model}-(b)), which expresses the reasoning steps as code and delegate computation steps to an code interpreter. It requires the reasoning steps to be expressed {\it accurately} as code. Therefore, we hypothesize that {\it prioritizing the learning of mathematical reasoning ability is helpful for the amplification of problem-solving ability}. In other words, {\it the initial learning with CoT} is essential for solving challenging mathematical problems, since it improves the mathematical reasoning ability~\cite{magister2022teaching, shridhar2023distilling, jie2023design, liang2023mint}. 

Our research is motivated by an analysis of existing models~\cite{tora, mammoth}. ToRA~\cite{tora} tried to learn reasoning ability as well as PoT by adding reasoning step into the PoT rationale. Similarly, MAmmoTH~\cite{mammoth} tried to learn both CoT and PoT by using both CoT rationale and PoT rationale as training data simultaneously. However, we conjecture that they do not fully utilize the advantages of learning with both CoT and PoT. This is because they did not consider {\it the sequence of CoT learning and PoT learning}, resulting a less effective learning.

In this work, we introduce a sequential learning approach, named \textsc{\textbf{{SAAS}}} (Solving Ability Amplification Strategy), to effectively utilize the strengths of CoT learning and PoT learning. This approach transitions from CoT learning to PoT learning, focusing on {\it enhancing problem-solving ability in PoT learning based on logical skills established in CoT learning}. This pedagogical strategy ensures that the competencies developed during CoT learning positively influence the PoT learning phase, leading to an overall improvement in solving challenging mathematical problems.

We validate the rationality and effectiveness of our \textsc{\textbf{{SAAS}}} via extensive experiments on the reputable benchmarks~\cite{gsm8k, math, gsm-hard, svamp, asdiv, tabmwp, mawps}. Most importantly, \textsc{\textbf{{SAAS}}} achieved state-of-the-art with remarkable performance. Through this, in this paper, we present a {\it novel and effective perspective} ({\it i.e.}, our hypothesis) within the field of mathematics.

\section{\textsc{\textbf{{SAAS}}}: Solving Ability Amplification Strategy}\label{sec:proposed_mehtod}

In this paper, we hypothesize that learning about the problem-solving ability is more effective after logical skills are well established. To explore this, we propose the sequential learning approach, named \textsc{\textbf{{SAAS}}} (Solving Ability Amplification Strategy), which transitions from CoT learning to PoT learning as shown in Figure~\ref{fig:model}. Our \textsc{\textbf{{SAAS}}} is motivated by the pedagogical strategy of human that first learns logical skills and then develops problem-solving abilities by solving numerous problems~\cite{glaser1984education}. In the following subsections, we describe CoT learning and PoT learning in details.

\subsection{Chain-of-Thought Learning}
It has been shown in various domains that CoT learning, which trains LLMs with data composed of CoT rationales, improves reasoning ability~\cite{jie2023design, liang2023mint}. Thus, we first fine-tune the LLM via CoT learning for improving mathematical reasoning ability. The primary objective in this phase is to optimize the model parameters for logically interpreting and responding to mathematical problems.

To achieve this, we employ a widely used optimization approach~\cite{metamath, tora} that seeks to find the optimal parameters, denoted as $\theta_{cot}^*$, which minimize the negative log-likelihood. This is expressed mathematically as:
\begin{equation}
    \small
    \arg\min_\theta -\frac{1}{|D_{cot}|}\sum_{(x_{cot}, y_{cot})\in D_{cot}}\log p_\theta(y_{cot} | x_{cot}),
\label{eq:1}
\end{equation}
where $\theta$ represents the learnable parameters of the LLM. The dataset $D_{cot}$ consists of ($x_{cot}$, $y_{cot}$) pairs, where 
$x_{cot}$ denotes a mathematical question, and $y_{cot}$ is the desired CoT rationale for that question.

This optimization process is designed to ensure that the model learns to generate CoT rationales that are logically consistent throughout the reasoning process. This is particularly important in the field of mathematics, since the rationale behind each step is as critical as the final answer. By minimizing the negative log-likelihood, we effectively guide the model to generate step-by-step explanations that mirror human problem-solving approaches, thus enhancing its overall reasoning capability.

This phase sets the foundation for the subsequent PoT learning phase, where the model's enhanced reasoning ability, developed through CoT training, is further refined and applied to more complex problem-solving scenarios.

\subsection{Program-of-Thought Learning}\label{subsec:pot}
Although the LLM optimized with parameters $\theta_{cot}^*$ demonstrates improved logical skills, it still exhibits limitations in problem-solving ability, particularly in computational accuracy~\cite{pot}, which will be empirically validated in section~\ref{appendix:RQ4}. To amplify this problem-solving ability, building upon the mathematical reasoning established in the CoT learning phase, we further fine-tune the LLM with $\theta_{cot}^*$ as its starting point using data composed of PoT rationales.

To accomplish this, we construct a dataset $D_{pot+cot}$ that consists of both PoT and CoT rationales. Notably, we integrate CoT rationales alongside PoT rationales in this dataset. This is because we observed that focusing exclusively on PoT rationales during this phase leads to a deterioration in mathematical reasoning ability in our experiments, as detailed in Table~\ref{tab:ablation}. To mitigate this {\it cognitive forgetting}, we introduce a {\it cognitive retention strategy}. This strategy involves randomly sampling CoT rationales and incorporating them into the PoT learning phase. Such a mixed approach ({\it i.e.}, congnitive retention strategy) ensures that the LLM retains its previously acquired reasoning skills while adapting to the new learning format.

The objective in this phase is to find the final optimal parameters $\theta^*$ of the LLM, which involves minimizing the following negative log-likelihood:
\begin{equation}
    \small
    \arg\min_{\theta_{cot}^*} -\frac{1}{|D_{pot+cot}|}\sum_{(x, y)\in D_{pot+cot}}\log p_{\theta_{cot}^*}(y | x),
\label{eq:1}
\end{equation}
where $x$ represents a mathematical question, and $y$ is the desired output, which could be either a PoT rationale or a CoT rationale, for the given question $x$. This approach aims to harmonize the strengths of both CoT and PoT learning, thereby equipping the LLM with enhanced computational accuracy and problem-solving abilities while maintaining its proficiency in logical reasoning.

\section{Experiments}\label{sec:experiments}
In this section, we conduct extensive experiments to answer the following key research questions (RQs):
\begin{itemize}[leftmargin=*, itemsep=0.1em]
\item \textbf{RQ1}: Does SAAS quantitatively outperform its competitors for solving challenging mathematical problems?
\item \textbf{RQ2}: Are two core strategies of SAAS (sequential learning, cognitive retention strategy) effective in improving the accuracy?
\item \textbf{RQ3}: Is SAAS effective in solving not only basic but also challenging mathematical problems?
\item \textbf{RQ4}: Does sequential learning that transitions from CoT learning to PoT learning help improve both the mathematical reasoning and computational accuracy?
\end{itemize}

\subsection{Experimental Settings}

\begin{figure*}[t!]
    \begin{center}
        \includegraphics[width=1.0\textwidth]{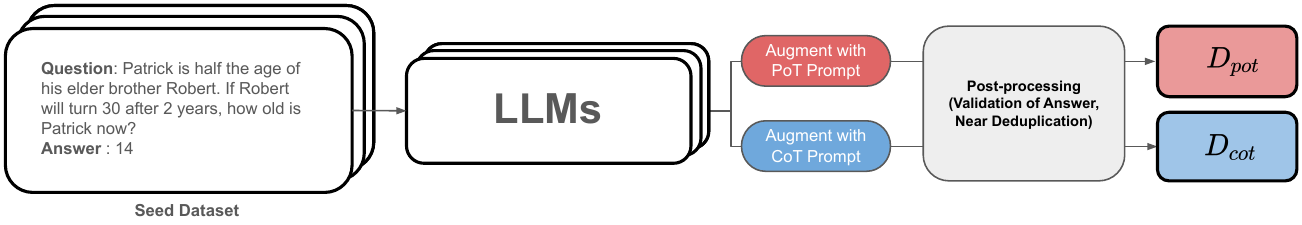}
    \end{center}
    \caption{Overall procedure of the synthetic data generation.}
    \label{fig:data_generation}
\end{figure*}

\begin{table}[t]
\centering
\footnotesize
\resizebox{0.48\textwidth}{!}{
\begin{tabular}{crrr}
\toprule
\textbf{Seed Dataset} & \textbf{Rationale} & \textbf{Models} & \textbf{Size} \\
\midrule
MetaMathQA & CoT & GPT, WizardMath & 465K \\
MATH, GSM8K & CoT & WizardMath & 300K \\
QANDA & CoT & WizardMath & 120K \\
\midrule
MetaMathQA & PoT & ToRA & 60K \\
MATH, GSM8K & PoT & ToRA & 226K \\
MathInstruct & PoT & ToRA & 38K \\
QANDA & PoT & ToRA & 12K \\
\bottomrule
\end{tabular}
}
\caption{Summary of synthetic datasets}
\label{tab:dataset}
\end{table}

\begin{table*}[h!]
\centering
    \resizebox{0.9\textwidth}{!}{
    \begin{tabular}{c|c|ccccccc|c}
    \toprule
    \textbf{Model} & \textbf{Size} & \textbf{GSM8K} & \textbf{MATH} & \textbf{GSM-Hard} & \textbf{SVAMP} & \textbf{TabMWP} & \textbf{ASDiv} & \textbf{MAWPS} & \textbf{Avg.} \\
    \midrule
    \multicolumn{10}{c}{\textbf{General Models}} \\
    \midrule
    \textbf{GPT-4} & - & 92.0 & 45.2 & 64.7 & 93.1 & 67.1 & 91.3 & 97.6 & 78.3 \\
    \textbf{GPT-4 (PAL)} & - & 94.2 & 51.8 & 77.6 & 94.8 & 95.9 & 92.6 & 97.7 & 86.4 \\
    \textbf{ChatGPT} & - & 80.8 & 35.5 & 55.9 & 83.0 & 69.1 & 87.3 & 94.6 & 72.3 \\
    \textbf{ChatGPT (PAL)} & - & 78.6 & 38.7 & 67.6 & 77.8 & 79.9 & 81.0 & 89.4 & 73.3 \\
    \textbf{Claude-2} & - & 85.2 & 32.5 & - & - & - & - & - & - \\
    \textbf{PaLM-2} & 540B & 80.7 & 34.3 & - & - & - & - & - & - \\
    \midrule
    \textbf{LLaMa-2} & 7B & 13.3 & 4.1 & 7.8 & 38.0 & 31.1 & 50.7 & 60.9 & 29.4 \\
    \textbf{Platypus-2} & 7B & 14.4 & 5.4 & 8.6 & 36.7 & 26.5 & 47.9 & 58.4 & 28.3 \\
    \textbf{CodeLLaMa (PAL)} & 7B & 34.0 & 16.6 & 33.6 & 59.0 & 47.3 & 61.4 & 79.6 & 47.4 \\
    \midrule
    \textbf{SOLAR-1} & 10.7B & 25.8 & 8.0 & 17.1 & 59.3 & 33.6 & 55.1 & 68.4 & 38.1 \\
    \textbf{LLaMa-2} & 13B & 24.3 & 6.3 & 13.6 & 43.1 & 39.5 & 56.3 & 70.4 & 36.2 \\
    \textbf{Platypus-2} & 13B & 23.7 & 7.1 & 14.3 & 50.7 & 45.3 & 55.1 & 69.6 & 38.0 \\
    \textbf{CodeLLaMa (PAL)} & 13B & 39.9 & 19.9 & 39.0 & 62.4 & 59.5 & 65.3 & 86.0 & 53.1 \\
    \midrule
    \textbf{CodeLLaMa (PAL)} & 34B & 53.3 & 23.9 & 49.4 & 71.0 & 63.1 & 72.4 & 91.5 & 60.7 \\
    \midrule
    \textbf{LLaMa-2} & 70B & 57.8 & 14.4 & 36.0 & 73.6 & 57.5 & 76.0 & 92.4 & 58.2 \\
    \textbf{Platypus-2} & 70B & 45.9 & 15.0 & 24.6 & 74.3 & 47.3 & 72.7 & 91.1 & 53.0 \\
    \midrule
    \multicolumn{10}{c}{\textbf{Mathematics Domain-Specific Models}} \\
    \midrule
    \textbf{WizardMath} & 7B & 54.9 & 10.7 & 20.6 & 57.3 & 38.1 & 59.1 & 73.7 & 44.9 \\
    \textbf{MetaMath} & 7B & 66.5 & 19.8 & - & - & - & - & - & - \\
    \textbf{MuggleMATH} & 7B & 68.4 & - & - & - & - & - & - & - \\
    \textbf{Toolformer} & 7B & - & - & - & 29.4 & - & 40.4 & 44.0 & - \\
    \textbf{MathCoder} & 7B & 64.2 & 23.3 & - & - & - & - & - & - \\
    \textbf{MathCoder-CODE} & 7B & 67.8 & 30.2 & - & - & - & - & - & - \\
    \textbf{MAmmoTH} & 7B & 53.6 & 31.5 & - & - & - & - & - & - \\
    \textbf{MAmmoTH-CODE} & 7B & 59.4 & 33.4 & - & - & - & - & - & - \\
    \textbf{ToRA} & 7B & 68.8 & 40.1 & 54.6 & 68.2 & 42.4 & 73.9 & 88.8 & 62.4 \\
    \cellcolor{lp!60}\textsc{\textbf{SAAS}} & \cellcolor{lp!60}7B & \cellcolor{lp!60}\underline{74.3} & \cellcolor{lp!60}\underline{43.2} & \cellcolor{lp!60}\textbf{58.3} & \cellcolor{lp!60}\textbf{74.3} & \cellcolor{lp!60}\underline{49.6} & \cellcolor{lp!60}\underline{77.3} & \cellcolor{lp!60}\underline{93.6} & \cellcolor{lp!60}\underline{67.2} \\
    \textbf{ToRA-CODE} & 7B & 72.6 & 44.6 & 56.0 & 70.4 & 51.6 & 78.7 & 91.3 & 66.5 \\
    \cellcolor{lp!60}\textsc{\textbf{{SAAS-CODE}}} & \cellcolor{lp!60}7B & \cellcolor{lp!60}\textbf{74.8} & \cellcolor{lp!60}\textbf{45.2} & \cellcolor{lp!60}\underline{58.1} & \cellcolor{lp!60}\underline{73.6} & \cellcolor{lp!60}\textbf{64.0} & \cellcolor{lp!60}\textbf{80.4} & \cellcolor{lp!60}\textbf{93.8} & \cellcolor{lp!60}\textbf{70.0} \\
    \midrule
    \cellcolor{lp!60}\textsc{\textbf{{SAAS}}} & \cellcolor{lp!60}10.7B & \cellcolor{lp!60}\textbf{82.0} & \cellcolor{lp!60}\underline{50.1} & \cellcolor{lp!60}\textbf{64.9} & \cellcolor{lp!60}\textbf{85.0} & \cellcolor{lp!60}\textbf{72.5} & \cellcolor{lp!60}\textbf{87.5} & \cellcolor{lp!60}\textbf{95.7} & \cellcolor{lp!60}\textbf{76.8} \\
    \textbf{WizardMath} & 13B & 63.9 & 14.0 & 28.4 & 64.3 & 46.7 & 65.8 & 79.7 & 51.8 \\
    \textbf{MetaMath} & 13B & 72.3 & 22.4 & - & - & - & - & - & - \\
    \textbf{MuggleMATH} & 13B & 74.0 & - & - & - & - & - & - & - \\
    \textbf{MathCoder} & 13B & 72.6 & 29.9 & - & - & - & - & - & - \\
    \textbf{MathCoder-CODE} & 13B & 74.1 & 35.9 & - & - & - & - & - & - \\
    \textbf{MAmmoTH} & 13B & 62.0 & 34.2 & - & - & - & - & - & - \\
    \textbf{MAmmoTH-CODE} & 13B & 64.7 & 36.3 & - & - & - & - & - & - \\
    \textbf{ToRA} & 13B & 72.7 & 43.0 & 57.3 & 72.9 & 47.2 & 77.2 & 91.3 & 65.9 \\
    \cellcolor{lp!60}\textsc{\textbf{{SAAS}}} & \cellcolor{lp!60}13B & \cellcolor{lp!60}76.6 & \cellcolor{lp!60}46.2 & \cellcolor{lp!60}\underline{61.6} & \cellcolor{lp!60}77.8 & \cellcolor{lp!60}58.2 & \cellcolor{lp!60}80.5 & \cellcolor{lp!60}94.3 & \cellcolor{lp!60}70.7 \\
    \textbf{ToRA-CODE} & 13B & 75.8 & 48.1 & 60.5 & 75.7 & 65.4 & 81.4 & 92.5 & 71.3 \\
    \cellcolor{lp!60}\textsc{\textbf{{SAAS-CODE}}} & \cellcolor{lp!60}13B & \cellcolor{lp!60}\underline{79.4} & \cellcolor{lp!60}\textbf{50.6} & \cellcolor{lp!60}\underline{61.6} & \cellcolor{lp!60}\underline{80.6} & \cellcolor{lp!60}\underline{68.2} & \cellcolor{lp!60}\underline{84.5} & \cellcolor{lp!60}\underline{95.4} & \cellcolor{lp!60}\underline{74.3} \\
    \midrule
    \textbf{MathCoder-CODE} & 34B & 81.7 & 45.2 & - & - & - & - & - & - \\
    \textbf{MAmmoTH-CODE} & 34B & 72.7 & 43.6 & - & - & - & - & - & - \\
    \textbf{ToRA-CODE} & 34B & {80.7} & {50.8} & {63.7} & {80.5} & {70.5} & {84.2} & {93.3} & {74.8} \\
    \cellcolor{lp!60}\textsc{\textbf{{SAAS-CODE}}} & \cellcolor{lp!60}34B & \cellcolor{lp!60}\underline{82.9} & \cellcolor{lp!60}\underline{52.3} & \cellcolor{lp!60}\underline{64.1} & \cellcolor{lp!60}\underline{82.8} & \cellcolor{lp!60}\underline{73.9} & \cellcolor{lp!60}\underline{85.4} & \cellcolor{lp!60}\underline{95.2} & \cellcolor{lp!60}\underline{76.6} \\
    \cellcolor{lp!60}\textsc{\textbf{{SAAS-Llema}}} & \cellcolor{lp!60}34B & \cellcolor{lp!60}\textbf{85.4} & \cellcolor{lp!60}\textbf{54.7} & \cellcolor{lp!60}\textbf{67.0} & \cellcolor{lp!60}\textbf{85.2} & \cellcolor{lp!60}\textbf{80.2} & \cellcolor{lp!60}\textbf{87.6} & \cellcolor{lp!60}\textbf{96.6} & \cellcolor{lp!60}\textbf{79.5} \\
    \midrule
    \textbf{WizardMath} & 70B & 81.6 & 22.7 & 50.3 & 80.0 & 49.8 & 76.2 & 86.2 & 63.8 \\
    \textbf{MetaMath} & 70B & 82.3 & 26.6 & - & - & - & - & - & - \\
    \textbf{MuggleMATH} & 70B & 82.3 & - & - & - & - & - & - & - \\
    \textbf{MathCoder} & 70B & 83.9 & 45.1 & - & - & - & - & - & - \\
    \textbf{ToRA} & 70B & 84.3 & 49.7 & 67.2 & 82.7 & 74.0 & 86.8 & 93.8 & 76.9 \\
\bottomrule
\end{tabular}
}
\caption{Accuracies of competitors and our \textsc{\textbf{{SAAS}}} on the mathematical benchmark datasets. Our \textsc{\textbf{{SAAS}}} models are shown in purple color.}
\label{tab:main}
\end{table*}

\subsubsection{Dataset Details}
In this paper, we synthesize GSM8K~\cite{gsm8k}, MATH~\cite{math}, MetaMathQA~\cite{metamath}, MathInstruct~\cite{mammoth}, and QANDA. The QANDA dataset was gathered manually through direct interaction with the application\footnote{https://mathpresso.com/en}. The overall procedure of synthetic data generation is illustrated in Figure~\ref{fig:data_generation}.

Specifically, we synthesize these datasets into Chain-of-Thought (CoT) and Program-of-Thought (PoT) rationales via various models (GPT, WizardMath~\cite{wizardmath}, ToRA~\cite{tora}). To generate diverse synthetic data, we adjust some hyperparameters such as temperature and top\_p. Then, we select only the correct responses and eliminate similar ones among these correct responses as in~\citet{self_instruct}. 
The detailed descriptions of seed datasets are described in Appendix~\ref{appendix:a}. Table~\ref{tab:dataset} provides the summary of our synthetic datasets for fine-tuning.

\subsubsection{Training Details}
We used the CodeLLaMA 13B model~\cite{codellama} as our base model and fine-tuned it with our synthetic datasets by setting the batch size to 128. We set learning rate to $2e-5$ and use cosine scheduler with warm-up period (1 epoch). For efficient model training, we used DeepSpeed ZeRO Stage3~\cite{deepspeed}.

\subsubsection{Model Details}

To evaluate the effectiveness of our SAAS in RQ1, we compared it with several state-of-the-art competitors. These competitors are divided into two groups: general models and mathematics domain-specific models. The general models include GPT-4~\cite{gpt4}, ChatGPT (gpt-3.5-turbo)\cite{chatgpt}, Claude-2\cite{claude2}, PaLM-2~\cite{palm2}, LLaMA-2~\cite{llama2}, Platypus-2~\cite{platypus2}, CodeLLaMA~\cite{codellama}, and SOLAR-1~\cite{solar}. The mathematics domain-specific models consist of WizardMath~\cite{wizardmath}, MetaMath~\cite{metamath}, MulggleMath~\cite{muggle}, Toolformer~\cite{toolformer}, MathCoder~\cite{mathcoder}, MammoTH~\cite{mammoth}, and ToRA~\cite{tora}.

As in~\citet{tora}, we report CoT prompting results by default, and include PAL~\cite{pal} prompting results for selected models. Within the category of mathematics domain-specific models, WizardMath, MetaMath, and MuggleMath exclusively employ CoT learning for fine-tuning. Conversely, ToRA utilizes solely PoT learning, whereas MathCoder and MammoTh integrate a combination of CoT and PoT learning methodologies for fine-tuning. Also, Toolformer is trained to utilize calculators.

\subsubsection{Evaluation Details}
We evaluated the model's performance and its ability to generalize mathematical reasoning using both in-domain and out-of-domain data. For in-domain evaluation, we use the test set of MATH and GSM8K dataset. For out-of-domain evaluation, we utilized the following various datasets, which are used in the previous studies~\cite{tora, mammoth} and publicly available: GSM-Hard~\cite{gsm-hard}, SVAMP~\cite{svamp}, ASDIV~\cite{asdiv}, TabMWP~\cite{tabmwp}, and MAWPS~\cite{mawps} that consists of SingleEQ, SingleOP, AddSub, and MultiArith. These datasets ensure a comprehensive analysis of the model's applicability across various mathematical contexts.

\subsection{Results and Analysis}\label{subsec:results}
We highlight the best and the second-best results in each column ({\it i.e.}, dataset) of the following tables in bold and underline, respectively.

\subsubsection{RQ1: Comparison with Competitors}
To demonstrate the superiority of our \textsc{\textbf{{SAAS}}} over competitors, we compare the accuracies of all competitors and \textsc{\textbf{{SAAS}}}. In this experiment, we utilize LLaMA-2 7B, CodeLLaMA 7B, SOLAR-1 10.7B, LLaMA-2 13B, CodeLLaMA 13B, CodeLLaMA 34B, and Llemma-34B as our base models.\footnote{For experiment on the 70B model, we could not proceed it due to hardware constraint.}

Table~\ref{tab:main} shows the results. We summarize our empirical findings as follows. First, we observed that mathematics domain-specific models outperforms general models {\it with similar size} in almost cases. This indicates a requisite for domain-specific models to address complex mathematical problems effectively. Second, among mathematics domain-specific competitors, ToRA, which utilizes solely PoT learning, {\it consistently} outperforms all others with similar size, including MathCoder and MammoTH, which integrate a combination of CoT learning and PoT learning methodologies. This implies that simply combining CoT and PoT learning does not effectively solve complex mathematical problems. Therefore, a strategic and careful approach is imperative in the combination of CoT and PoT learning. Third and most importantly, our \textsc{\textbf{SAAS}} {\it consistently} and {\it significantly} outperforms all competitors with similar size. Specifically, on $\sim$7B size, 7B$\sim$13B size, 13B$\sim$34B size, and 34B$\sim$70B size, \textsc{\textbf{SAAS}} outperforms the best competitors ({\it i.e.}, ToRA-CODE and ToRA) by up to 5.26\%, 7.71\%, and 6.28\% in terms of average score.
Note that although we could not fine-tune 70B model, \textsc{\textbf{SAAS}} with 10.7B showed similar performance to ToRA with 70B. Furthermore, \textsc{\textbf{{SAAS-Llema}}} demonstrated superior performance than ToRA with 70B. This remarkable performance of \textsc{\textbf{SAAS}} underscore the effectiveness of our sequential learning approach.

\begin{table}[t]
\centering
\resizebox{0.48\textwidth}{!}{
\begin{tabular}{lcc}
\toprule
\textbf{Strategy} & \textbf{GSM8K} & \textbf{MATH} \\
\midrule
Chain-of-Thought (CoT) & 69.7 & 26.9 \\
Program-of-Thought (PoT) & 76.8 & 47.7 \\
Combination of CoT and PoT & \underline{79.0} & 49.2 \\
\textsc{\textbf{{SAAS}}} & \textbf{79.4} & \textbf{50.6} \\
~~without cognitive retention strategy & \underline{79.0} & \underline{49.6} \\
Reverse \textsc{\textbf{{SAAS}}} & 76.8 & 47.1 \\
~~without cognitive retention strategy & 69.4 & 27.6 \\
\bottomrule
\end{tabular}
}
\caption{Accuracies of different learning strategies. All improvements are statistically significant with $p$-value $\leq 0.001$.}
\label{tab:ablation}
\end{table}

\subsubsection{RQ2: Effectiveness of Sequential Learning and Cognitive Retention Strategy}
To further explore what factors contribute to the improvement of our \textsc{\textbf{{SAAS}}}, we conduct comparative experiments on diverse learning strategies, as shown in Table~\ref{tab:ablation}. Specifically, we compare CoT learning, PoT learning, CoT+PoT learning, \textsc{\textbf{{SAAS}}} that transtions from CoT learning to PoT learning, and reverse \textsc{\textbf{{SAAS}}} that transtions from PoT learning to CoT learning. In addition, we compare (reverse) \textsc{\textbf{{SAAS}}} {\it without cognitive retention strategy} to validate the effectiveness of this strategy.
From Table~\ref{tab:ablation}, our empirical findings are summarized as follows:
\begin{longenum}
\item \textbf{Effectiveness of the hybrid learning}: Combining of CoT and PoT learning significantly outperforms both CoT learning and PoT learning. 
This is because CoT learning, which enhances mathematical reasoning ability, and PoT learning, which improves problem-solving ability, play a complementary role;
\item \textbf{Effectiveness of the sequential learning}: Our \textsc{\textbf{{SAAS}}} without cognitive retention strategy {\it slightly} outperforms combining of CoT and PoT learning in MATH only. We conjecture that the absence of significant improvement, despite sequential learning, can be attributed to the deterioration of mathematical reasoning abilities during the PoT learning phase ({\it i.e.}, cognitive forgetting). Furthermore, reverse \textsc{\textbf{{SAAS}}} without cognitive retention strategy shows a lower accuracy than combining of CoT and PoT learning. This result indicates that the order of the learning sequences in sequential learning is vital for mathematical reasoning and problem-solving abilities;
\item \textbf{Effectiveness of the cognitive retention strategy}: To mitigate the cognitive forgetting, in Section~\ref{subsec:pot}, we proposed the cognitive retention strategy, which includes some data samples from first phase in the second phase. (Reverse) \textsc{\textbf{{SAAS}}} outperforms (reverse) \textsc{\textbf{{SAAS}}} without cognitive retention strategy, verifying the effectiveness of the cognitive retention strategy.
\end{longenum}

\subsubsection{RQ3: Further Analysis of the Capabilities of \textsc{\textbf{{SAAS}}}}
\label{appendix:RQ3}
To analyze the capabilities of \textsc{\textbf{{SAAS}}} depending on the difficuly of mathemtical problem, we quantitatively assess the break-down accuracies for problems with respect to the reasoning steps as in~\citet{shi2023large}. Specifically, we segmented the GSM8K dataset into 4 categories based on the number of reasoning steps required to arrive at an answer. Then, we quantified accuracies of CoT learning, PoT learning, and SAAS across each designated category.

\begin{figure}
    \centering
    \includegraphics[width=0.48\textwidth]{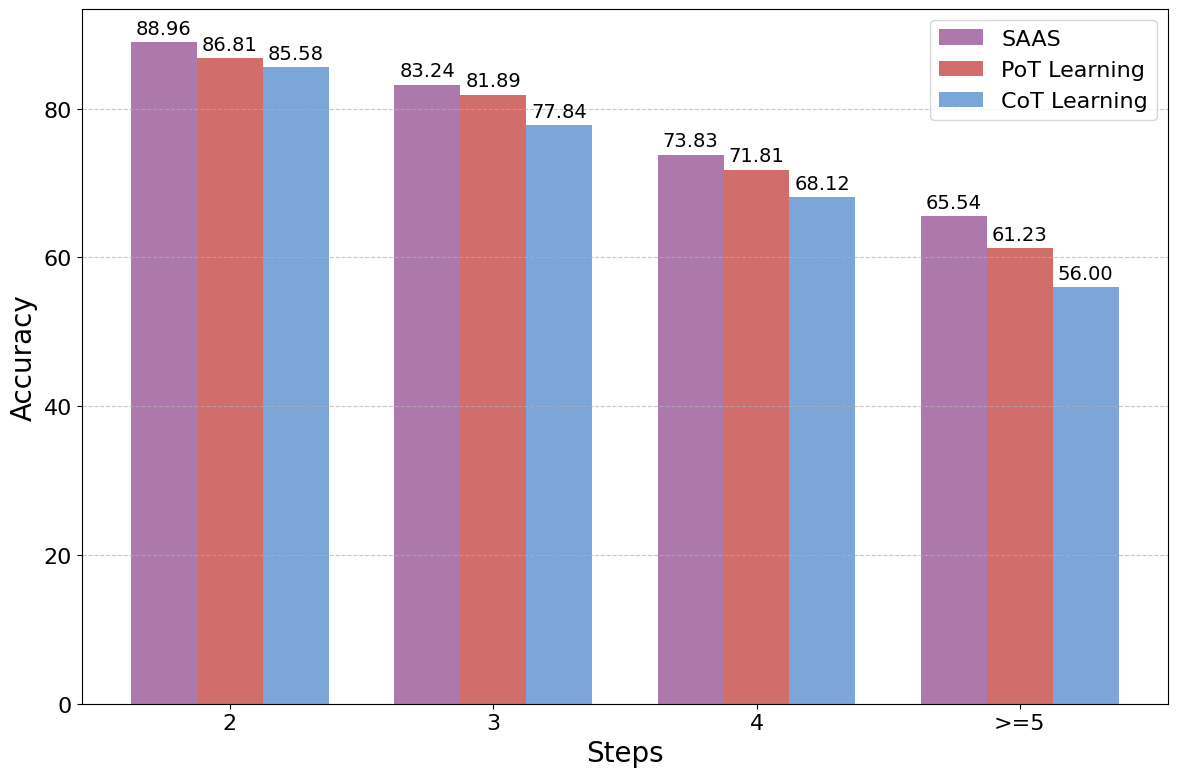}
    \caption{Accuracies on GSM8K with respect to the number of required reasoning steps.}
    \label{fig:enter-label}
\end{figure}

\begin{figure*}[t!]
    \centering
    \includegraphics[width=1.0\textwidth]{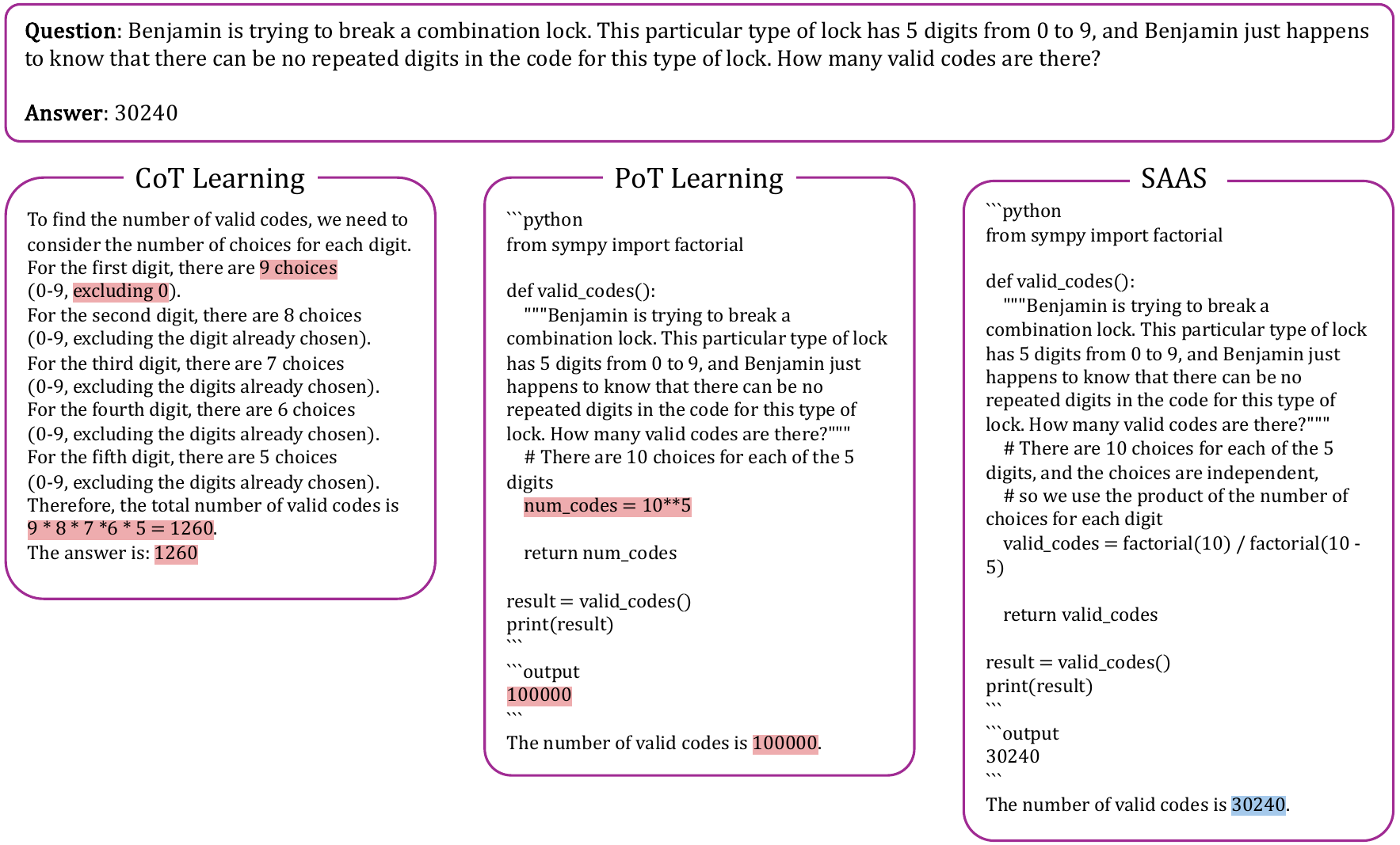}
    \caption{Responses of different learning approaches for a given question-answer pair.}
    \label{fig:case_study}
\end{figure*}

As illustrated in Figure~\ref{fig:enter-label}, \textsc{\textbf{{SAAS}}} outperforms both CoT learning and PoT learning for all categories. Moreover, as the reasoning steps in a mathematical problem extend ({\it i.e.}, the difficulty increases), especially the step 5 or above, the difference between our \textsc{\textbf{{SAAS}}} and other strategies becomes more pronounced. This result supports our hypothesis that prioritizing the learning of mathematical reasoning ability via CoT learning is helpful for the amplification of {\it challenging} problem-solving ability.

\subsubsection{RQ4: Case Study}\label{appendix:RQ4}
To demonstrate that our \textsc{\textbf{SAAS}} is effective in terms of both mathematical reasoning and computational accuracy, we conduct a case study showing the responses of CoT learning, PoT learning, and \textsc{\textbf{SAAS}} for a given question-answer pair.
Figure~\ref{fig:case_study} shows the visualization results, where the colored words indicate incorrect responses and the words with no color mark indicate correct responses.

As depicted in Figure~\ref{fig:case_study}, CoT learning approach exhibited inaccuracies in arithmetic computations as well as deficiencies in mathematical reasoning. Conversely, PoT approach demonstrated precise calculations yet exhibited a critical deficiency in mathematical reasoning. As we expected, our \textsc{\textbf{SAAS}} exhibited precise computational accuracy along with enhanced mathematical reasoning capabilities (See the more detailed comments than the comments of PoT learning). Through this case study, we demonstrated the following three observations: i) only CoT learning approach leads to arithmetic calculation errors; ii) only PoT learning approach may result in a deficit of mathematical reasoning; iii) sequential learning that transitions from CoT to PoT learning help improve computational accuracy as well as mathematical reasoning.

\section{Conclusion}
In this paper, we demonstrated the following two important points in the sense of solving challenging mathematical problems: (1) prioritizing the learning of mathematical reasoning ability via Chain-of-Thought (CoT) learning is helpful for the amplification of problem-solving ability during Program-of-Thought (PoT) learning; (2) for effective sequential learning, it is necessary to employ {\it a cognitive retention strategy} that incorporates some data samples from the initial phase into the subsequent phase. In light of this, we proposed a novel sequential learning approach, named \textsc{\textbf{SAAS}} (Solving Ability Amplification Strategy), which progresses from CoT learning to PoT learning with cognitive retention strategy.
Through extensive experiments with the reputable benchmarks, we demonstrated that \textsc{\textbf{SAAS}} consistently and significantly outperforms all competitor, marking a significant advancement in the field of mathematical reasoning in LLMs.

\section*{Acknowledgements}
This work was supported by the 2023 KT ICT AI2XL Laboratory R\&D Fund" project funded by KT. 

\section*{Limitations}
This study, while advancing the field of computational linguistics through the use of Large Language Models (LLMs), encounters several limitations that are important to acknowledge.

Firstly, the intricate nature of LLMs can sometimes lead to unpredictability in their outputs. This unpredictability can be particularly challenging when dealing with mathematical reasoning, where precision and accuracy are paramount, making it difficult to utilize LLMs in applications in the field of mathematics.

Furthermore, despite advancements via our study, LLMs still have limitations in their understanding and application of advanced mathematical concepts. While they can perform well on structured problems, their ability to handle abstract and complex mathematical reasoning is still an area of ongoing research and development.

Additionally, the reliance on synthetic data for training these models also presents a limitation. While synthetic datasets are useful for mitigating the scarcity of real-world data, it may not always accurately capture real-world scenarios, leading to potential gaps in the model’s performance when applied to practical, real-world tasks.

Finally, ethical considerations, particularly around the potential misuse of AI, remain a concern. Ensuring that LLMs are used responsibly and do not perpetuate biases is an ongoing challenge in the field.

In summary, while our study leverages the capabilities of LLMs to enhance mathematical reasoning in computational linguistics, it is important to recognize the limitations related to unpredictability of LLMs, understanding of advanced mathematical concepts, reliance on synthetic data, and ethical considerations. These limitations highlight the need for continued research and development in the field to address these challenges effectively.

\section*{Ethics Statement}
In this research, we have diligently adhered to the highest ethical standards of scientific inquiry and data management, ensuring the integrity and reliability of our findings. The design and execution of our experiments were grounded in fairness and objectivity, without favoring any particular outcome. This commitment was reflected in our meticulous planning and consistent application of methodologies across various datasets. 

We also placed a strong emphasis on data privacy and security, handling all data, especially synthetic data generated for our models, in compliance with relevant data protection laws and guidelines. We confirmed that all the data used in our experiments were free of licensing issues. Our approach to data was characterized by strict anonymization protocols and its use was confined strictly to research purposes. We have strived for transparency in our research process, documenting all methodologies, data sources, and analysis techniques clearly, which underpins our commitment to the reproducibility of scientific research. This allows other researchers to verify our results and build upon our work, contributing to the collective knowledge in the field. 

Recognizing the broader impacts of AI and LLMs on society, our research was conducted with a profound sense of responsibility. We were mindful of the ethical implications of AI development and aimed to create models that are effective yet ethically aligned, avoiding any form of biased, discriminatory, or harmful applications of these technologies. We believe our research makes a positive contribution to the field of computational linguistics and AI, particularly in enhancing the mathematical reasoning capabilities of Large Language Models in a manner that is ethically sound and socially responsible. 

Our work underscores our commitment to conducting scientifically rigorous and ethically responsible research, maintaining the highest standards of integrity in AI and computational linguistics.

\bibliography{anthology,custom}

\begin{thebibliography}{54}
\expandafter\ifx\csname natexlab\endcsname\relax\def\natexlab#1{#1}\fi

\bibitem[{Achiam et~al.(2023)Achiam, Adler, Agarwal, Ahmad, Akkaya, Aleman, Almeida, Altenschmidt, Altman, Anadkat et~al.}]{gpt4}
Josh Achiam, Steven Adler, Sandhini Agarwal, Lama Ahmad, Ilge Akkaya, Florencia~Leoni Aleman, Diogo Almeida, Janko Altenschmidt, Sam Altman, Shyamal Anadkat, et~al. 2023.
\newblock Gpt-4 technical report.
\newblock \emph{arXiv preprint arXiv:2303.08774}.

\bibitem[{Amini et~al.(2019)Amini, Gabriel, Lin, Koncel-Kedziorski, Choi, and Hajishirzi}]{mathqa}
Aida Amini, Saadia Gabriel, Peter Lin, Rik Koncel-Kedziorski, Yejin Choi, and Hannaneh Hajishirzi. 2019.
\newblock Mathqa: Towards interpretable math word problem solving with operation-based formalisms.
\newblock \emph{arXiv preprint arXiv:1905.13319}.

\bibitem[{Anil et~al.(2023)Anil, Dai, Firat, Johnson, Lepikhin, Passos, Shakeri, Taropa, Bailey, Chen et~al.}]{palm2}
Rohan Anil, Andrew~M Dai, Orhan Firat, Melvin Johnson, Dmitry Lepikhin, Alexandre Passos, Siamak Shakeri, Emanuel Taropa, Paige Bailey, Zhifeng Chen, et~al. 2023.
\newblock Palm 2 technical report.
\newblock \emph{arXiv preprint arXiv:2305.10403}.

\bibitem[{{Anthropic}(2023)}]{claude2}
{Anthropic}. 2023.
\newblock Model card and evaluations for claude models.
\newblock URL \url{https://www-files.anthropic.com/production/images/Model-Card-Claude-2.pdf}.

\bibitem[{Azerbayev et~al.(2023)Azerbayev, Schoelkopf, Paster, Santos, McAleer, Jiang, Deng, Biderman, and Welleck}]{llemma}
Zhangir Azerbayev, Hailey Schoelkopf, Keiran Paster, Marco~Dos Santos, Stephen McAleer, Albert~Q Jiang, Jia Deng, Stella Biderman, and Sean Welleck. 2023.
\newblock Llemma: An open language model for mathematics.
\newblock \emph{arXiv preprint arXiv:2310.10631}.

\bibitem[{Chen et~al.(2022)Chen, Ma, Wang, and Cohen}]{pot}
Wenhu Chen, Xueguang Ma, Xinyi Wang, and William~W Cohen. 2022.
\newblock Program of thoughts prompting: Disentangling computation from reasoning for numerical reasoning tasks.
\newblock \emph{arXiv preprint arXiv:2211.12588}.

\bibitem[{Chen et~al.(2023)Chen, Yin, Ku, Lu, Wan, Ma, Xu, Wang, and Xia}]{theoremqa}
Wenhu Chen, Ming Yin, Max Ku, Pan Lu, Yixin Wan, Xueguang Ma, Jianyu Xu, Xinyi Wang, and Tony Xia. 2023.
\newblock Theoremqa: A theorem-driven question answering dataset.
\newblock \emph{arXiv preprint arXiv:2305.12524}.

\bibitem[{Cobbe et~al.(2021)Cobbe, Kosaraju, Bavarian, Chen, Jun, Kaiser, Plappert, Tworek, Hilton, Nakano et~al.}]{gsm8k}
Karl Cobbe, Vineet Kosaraju, Mohammad Bavarian, Mark Chen, Heewoo Jun, Lukasz Kaiser, Matthias Plappert, Jerry Tworek, Jacob Hilton, Reiichiro Nakano, et~al. 2021.
\newblock Training verifiers to solve math word problems.
\newblock \emph{arXiv preprint arXiv:2110.14168}.

\bibitem[{Dong et~al.(2023)Dong, Yuan, Lu, Li, Xue, Liu, Wang, Yuan, Zhou, and Zhou}]{dong2023abilities}
Guanting Dong, Hongyi Yuan, Keming Lu, Chengpeng Li, Mingfeng Xue, Dayiheng Liu, Wei Wang, Zheng Yuan, Chang Zhou, and Jingren Zhou. 2023.
\newblock How abilities in large language models are affected by supervised fine-tuning data composition.
\newblock \emph{arXiv preprint arXiv:2310.05492}.

\bibitem[{Drori et~al.(2021)Drori, Tran, Wang, Cheng, Liu, Tang, Ke, Singh, Patti, Lynch et~al.}]{drori2021neural}
Iddo Drori, Sunny Tran, Roman Wang, Newman Cheng, Kevin Liu, Leonard Tang, Elizabeth Ke, Nikhil Singh, Taylor~L Patti, Jayson Lynch, et~al. 2021.
\newblock A neural network solves and generates mathematics problems by program synthesis: Calculus, differential equations, linear algebra, and more.
\newblock \emph{CoRR, abs/2112.15594}.

\bibitem[{Gao et~al.(2023{\natexlab{a}})Gao, Madaan, Zhou, Alon, Liu, Yang, Callan, and Neubig}]{pal}
Luyu Gao, Aman Madaan, Shuyan Zhou, Uri Alon, Pengfei Liu, Yiming Yang, Jamie Callan, and Graham Neubig. 2023{\natexlab{a}}.
\newblock Pal: Program-aided language models.
\newblock In \emph{International Conference on Machine Learning}.

\bibitem[{Gao et~al.(2023{\natexlab{b}})Gao, Madaan, Zhou, Alon, Liu, Yang, Callan, and Neubig}]{gsm-hard}
Luyu Gao, Aman Madaan, Shuyan Zhou, Uri Alon, Pengfei Liu, Yiming Yang, Jamie Callan, and Graham Neubig. 2023{\natexlab{b}}.
\newblock Pal: Program-aided language models.
\newblock In \emph{International Conference on Machine Learning}, pages 10764--10799. PMLR.

\bibitem[{Glaser(1984)}]{glaser1984education}
Robert Glaser. 1984.
\newblock Education and thinking: The role of knowledge.
\newblock \emph{American psychologist}.

\bibitem[{Gou et~al.(2023)Gou, Shao, Gong, Yang, Huang, Duan, Chen et~al.}]{tora}
Zhibin Gou, Zhihong Shao, Yeyun Gong, Yujiu Yang, Minlie Huang, Nan Duan, Weizhu Chen, et~al. 2023.
\newblock Tora: A tool-integrated reasoning agent for mathematical problem solving.
\newblock \emph{arXiv preprint arXiv:2309.17452}.

\bibitem[{Hendrycks et~al.(2021)Hendrycks, Burns, Kadavath, Arora, Basart, Tang, Song, and Steinhardt}]{math}
Dan Hendrycks, Collin Burns, Saurav Kadavath, Akul Arora, Steven Basart, Eric Tang, Dawn Song, and Jacob Steinhardt. 2021.
\newblock Measuring mathematical problem solving with the math dataset.
\newblock \emph{arXiv preprint arXiv:2103.03874}.

\bibitem[{Jiang et~al.(2023)Jiang, Shi, Yu, Liu, Zhang, Li, and Kwok}]{fobar}
Weisen Jiang, Han Shi, Longhui Yu, Zhengying Liu, Yu~Zhang, Zhenguo Li, and James~T Kwok. 2023.
\newblock Forward-backward reasoning in large language models for mathematical verification.
\newblock \emph{arXiv preprint arXiv:2308.07758}.

\bibitem[{Jie et~al.(2023)Jie, Luong, Zhang, Jin, and Li}]{jie2023design}
Zhanming Jie, Trung~Quoc Luong, Xinbo Zhang, Xiaoran Jin, and Hang Li. 2023.
\newblock Design of chain-of-thought in math problem solving.
\newblock \emph{arXiv preprint arXiv:2309.11054}.

\bibitem[{Kim et~al.(2023)Kim, Park, Kim, Lee, Song, Kim, Kim, Kim, Lee, Kim et~al.}]{solar}
Dahyun Kim, Chanjun Park, Sanghoon Kim, Wonsung Lee, Wonho Song, Yunsu Kim, Hyeonwoo Kim, Yungi Kim, Hyeonju Lee, Jihoo Kim, et~al. 2023.
\newblock Solar 10.7 b: Scaling large language models with simple yet effective depth up-scaling.
\newblock \emph{arXiv preprint arXiv:2312.15166}.

\bibitem[{Koncel-Kedziorski et~al.(2016)Koncel-Kedziorski, Roy, Amini, Kushman, and Hajishirzi}]{mawps}
Rik Koncel-Kedziorski, Subhro Roy, Aida Amini, Nate Kushman, and Hannaneh Hajishirzi. 2016.
\newblock Mawps: A math word problem repository.
\newblock In \emph{Proceedings of the 2016 conference of the north american chapter of the association for computational linguistics: human language technologies}, pages 1152--1157.

\bibitem[{Lee et~al.(2023)Lee, Hunter, and Ruiz}]{platypus2}
Ariel~N Lee, Cole~J Hunter, and Nataniel Ruiz. 2023.
\newblock Platypus: Quick, cheap, and powerful refinement of llms.
\newblock \emph{arXiv preprint arXiv:2308.07317}.

\bibitem[{Lewkowycz et~al.(2022)Lewkowycz, Andreassen, Dohan, Dyer, Michalewski, Ramasesh, Slone, Anil, Schlag, Gutman-Solo et~al.}]{minerva}
Aitor Lewkowycz, Anders Andreassen, David Dohan, Ethan Dyer, Henryk Michalewski, Vinay Ramasesh, Ambrose Slone, Cem Anil, Imanol Schlag, Theo Gutman-Solo, et~al. 2022.
\newblock Solving quantitative reasoning problems with language models.
\newblock \emph{Advances in Neural Information Processing Systems}, 35:3843--3857.

\bibitem[{Li et~al.(2023{\natexlab{a}})Li, Yuan, Dong, Lu, Wu, Tan, Wang, and Zhou}]{muggle}
Chengpeng Li, Zheng Yuan, Guanting Dong, Keming Lu, Jiancan Wu, Chuanqi Tan, Xiang Wang, and Chang Zhou. 2023{\natexlab{a}}.
\newblock Query and response augmentation cannot help out-of-domain math reasoning generalization.
\newblock \emph{arXiv preprint arXiv:2310.05506}.

\bibitem[{Li et~al.(2023{\natexlab{b}})Li, Hammoud, Itani, Khizbullin, and Ghanem}]{camel-math}
Guohao Li, Hasan Abed Al~Kader Hammoud, Hani Itani, Dmitrii Khizbullin, and Bernard Ghanem. 2023{\natexlab{b}}.
\newblock Camel: Communicative agents for" mind" exploration of large scale language model society.
\newblock \emph{arXiv preprint arXiv:2303.17760}.

\bibitem[{Liang et~al.(2023)Liang, Yu, Pan, Yao, Zeng, Zhang, and Yu}]{liang2023mint}
Zhenwen Liang, Dian Yu, Xiaoman Pan, Wenlin Yao, Qingkai Zeng, Xiangliang Zhang, and Dong Yu. 2023.
\newblock Mint: Boosting generalization in mathematical reasoning via multi-view fine-tuning.
\newblock \emph{arXiv preprint arXiv:2307.07951}.

\bibitem[{Lightman et~al.(2023)Lightman, Kosaraju, Burda, Edwards, Baker, Lee, Leike, Schulman, Sutskever, and Cobbe}]{verify}
Hunter Lightman, Vineet Kosaraju, Yura Burda, Harri Edwards, Bowen Baker, Teddy Lee, Jan Leike, John Schulman, Ilya Sutskever, and Karl Cobbe. 2023.
\newblock Let's verify step by step.
\newblock \emph{arXiv preprint arXiv:2305.20050}.

\bibitem[{Ling et~al.(2017)Ling, Yogatama, Dyer, and Blunsom}]{aqua-rat}
Wang Ling, Dani Yogatama, Chris Dyer, and Phil Blunsom. 2017.
\newblock Program induction by rationale generation: Learning to solve and explain algebraic word problems.
\newblock \emph{arXiv preprint arXiv:1705.04146}.

\bibitem[{Lu et~al.(2022{\natexlab{a}})Lu, Qiu, Chang, Wu, Zhu, Rajpurohit, Clark, and Kalyan}]{tabmwp}
Pan Lu, Liang Qiu, Kai-Wei Chang, Ying~Nian Wu, Song-Chun Zhu, Tanmay Rajpurohit, Peter Clark, and Ashwin Kalyan. 2022{\natexlab{a}}.
\newblock Dynamic prompt learning via policy gradient for semi-structured mathematical reasoning.
\newblock \emph{arXiv preprint arXiv:2209.14610}.

\bibitem[{Lu et~al.(2022{\natexlab{b}})Lu, Qiu, Yu, Welleck, and Chang}]{survey_math}
Pan Lu, Liang Qiu, Wenhao Yu, Sean Welleck, and Kai-Wei Chang. 2022{\natexlab{b}}.
\newblock A survey of deep learning for mathematical reasoning.
\newblock \emph{arXiv preprint arXiv:2212.10535}.

\bibitem[{Luo et~al.(2023)Luo, Sun, Xu, Zhao, Lou, Tao, Geng, Lin, Chen, and Zhang}]{wizardmath}
Haipeng Luo, Qingfeng Sun, Can Xu, Pu~Zhao, Jianguang Lou, Chongyang Tao, Xiubo Geng, Qingwei Lin, Shifeng Chen, and Dongmei Zhang. 2023.
\newblock Wizardmath: Empowering mathematical reasoning for large language models via reinforced evol-instruct.
\newblock \emph{arXiv preprint arXiv:2308.09583}.

\bibitem[{Magister et~al.(2022)Magister, Mallinson, Adamek, Malmi, and Severyn}]{magister2022teaching}
Lucie~Charlotte Magister, Jonathan Mallinson, Jakub Adamek, Eric Malmi, and Aliaksei Severyn. 2022.
\newblock Teaching small language models to reason.
\newblock \emph{arXiv preprint arXiv:2212.08410}.

\bibitem[{Meadows and Freitas(2022)}]{survey_mlp}
Jordan Meadows and Andr{\'e} Freitas. 2022.
\newblock A survey in mathematical language processing.
\newblock \emph{arXiv preprint arXiv:2205.15231}.

\bibitem[{Miao et~al.(2021)Miao, Liang, and Su}]{asdiv}
Shen-Yun Miao, Chao-Chun Liang, and Keh-Yih Su. 2021.
\newblock A diverse corpus for evaluating and developing english math word problem solvers.
\newblock \emph{arXiv preprint arXiv:2106.15772}.

\bibitem[{Mishra et~al.(2022)Mishra, Mitra, Varshney, Sachdeva, Clark, Baral, and Kalyan}]{numglue}
Swaroop Mishra, Arindam Mitra, Neeraj Varshney, Bhavdeep Sachdeva, Peter Clark, Chitta Baral, and Ashwin Kalyan. 2022.
\newblock Numglue: A suite of fundamental yet challenging mathematical reasoning tasks.
\newblock \emph{arXiv preprint arXiv:2204.05660}.

\bibitem[{{OpenAI}(2023)}]{chatgpt}
{OpenAI}. 2023.
\newblock Chat-gpt.
\newblock URL \url{https://openai.com/blog/chatgpt}.

\bibitem[{Patel et~al.(2021)Patel, Bhattamishra, and Goyal}]{svamp}
Arkil Patel, Satwik Bhattamishra, and Navin Goyal. 2021.
\newblock Are nlp models really able to solve simple math word problems?
\newblock \emph{arXiv preprint arXiv:2103.07191}.

\bibitem[{Qian et~al.(2022)Qian, Wang, Li, Li, and Yan}]{qian2022limitations}
Jing Qian, Hong Wang, Zekun Li, Shiyang Li, and Xifeng Yan. 2022.
\newblock Limitations of language models in arithmetic and symbolic induction.
\newblock \emph{arXiv preprint arXiv:2208.05051}.

\bibitem[{Rajbhandari et~al.(2020)Rajbhandari, Rasley, Ruwase, and He}]{deepspeed}
Samyam Rajbhandari, Jeff Rasley, Olatunji Ruwase, and Yuxiong He. 2020.
\newblock Zero: Memory optimizations toward training trillion parameter models.
\newblock In \emph{SC20: International Conference for High Performance Computing, Networking, Storage and Analysis}.

\bibitem[{Roziere et~al.(2023)Roziere, Gehring, Gloeckle, Sootla, Gat, Tan, Adi, Liu, Remez, Rapin et~al.}]{codellama}
Baptiste Roziere, Jonas Gehring, Fabian Gloeckle, Sten Sootla, Itai Gat, Xiaoqing~Ellen Tan, Yossi Adi, Jingyu Liu, Tal Remez, J{\'e}r{\'e}my Rapin, et~al. 2023.
\newblock Code llama: Open foundation models for code.
\newblock \emph{arXiv preprint arXiv:2308.12950}.

\bibitem[{Schick et~al.(2023)Schick, Dwivedi-Yu, Dess{\i}, Raileanu, Lomeli, Zettlemoyer, Cancedda, and Scialom}]{toolformer}
Timo Schick, Jane Dwivedi-Yu, R~Dess{\i}, Roberta Raileanu, Maria Lomeli, Luke Zettlemoyer, Nicola Cancedda, and Thomas Scialom. 2023.
\newblock Toolformer: Language models can teach themselves to use tools (2023).
\newblock \emph{arXiv preprint arXiv:2302.04761}.

\bibitem[{Shi et~al.(2023)Shi, Chen, Misra, Scales, Dohan, Chi, Sch{\"a}rli, and Zhou}]{shi2023large}
Freda Shi, Xinyun Chen, Kanishka Misra, Nathan Scales, David Dohan, Ed~H Chi, Nathanael Sch{\"a}rli, and Denny Zhou. 2023.
\newblock Large language models can be easily distracted by irrelevant context.
\newblock In \emph{International Conference on Machine Learning}, pages 31210--31227. PMLR.

\bibitem[{Shridhar et~al.(2023)Shridhar, Stolfo, and Sachan}]{shridhar2023distilling}
Kumar Shridhar, Alessandro Stolfo, and Mrinmaya Sachan. 2023.
\newblock Distilling reasoning capabilities into smaller language models.
\newblock In \emph{Findings of the Association for Computational Linguistics: ACL 2023}, pages 7059--7073.

\bibitem[{Thawani et~al.(2021)Thawani, Pujara, Szekely, and Ilievski}]{survey_representing}
Avijit Thawani, Jay Pujara, Pedro~A Szekely, and Filip Ilievski. 2021.
\newblock Representing numbers in nlp: a survey and a vision.
\newblock \emph{arXiv preprint arXiv:2103.13136}.

\bibitem[{Touvron et~al.(2023)Touvron, Martin, Stone, Albert, Almahairi, Babaei, Bashlykov, Batra, Bhargava, Bhosale et~al.}]{llama2}
Hugo Touvron, Louis Martin, Kevin Stone, Peter Albert, Amjad Almahairi, Yasmine Babaei, Nikolay Bashlykov, Soumya Batra, Prajjwal Bhargava, Shruti Bhosale, et~al. 2023.
\newblock Llama 2: Open foundation and fine-tuned chat models.
\newblock \emph{arXiv preprint arXiv:2307.09288}.

\bibitem[{Wang et~al.(2023)Wang, Ren, Zhou, Lu, Luo, Shi, Zhang, Song, Zhan, and Li}]{mathcoder}
Ke~Wang, Houxing Ren, Aojun Zhou, Zimu Lu, Sichun Luo, Weikang Shi, Renrui Zhang, Linqi Song, Mingjie Zhan, and Hongsheng Li. 2023.
\newblock Mathcoder: Seamless code integration in llms for enhanced mathematical reasoning.
\newblock \emph{arXiv preprint arXiv:2310.03731}.

\bibitem[{Wang et~al.(2022)Wang, Kordi, Mishra, Liu, Smith, Khashabi, and Hajishirzi}]{self_instruct}
Yizhong Wang, Yeganeh Kordi, Swaroop Mishra, Alisa Liu, Noah~A Smith, Daniel Khashabi, and Hannaneh Hajishirzi. 2022.
\newblock Self-instruct: Aligning language model with self generated instructions.
\newblock \emph{arXiv preprint arXiv:2212.10560}.

\bibitem[{Wei et~al.(2022{\natexlab{a}})Wei, Tay, Bommasani, Raffel, Zoph, Borgeaud, Yogatama, Bosma, Zhou, Metzler et~al.}]{wei2022emergent}
Jason Wei, Yi~Tay, Rishi Bommasani, Colin Raffel, Barret Zoph, Sebastian Borgeaud, Dani Yogatama, Maarten Bosma, Denny Zhou, Donald Metzler, et~al. 2022{\natexlab{a}}.
\newblock Emergent abilities of large language models.
\newblock \emph{arXiv preprint arXiv:2206.07682}.

\bibitem[{Wei et~al.(2022{\natexlab{b}})Wei, Wang, Schuurmans, Bosma, Xia, Chi, Le, Zhou et~al.}]{cot}
Jason Wei, Xuezhi Wang, Dale Schuurmans, Maarten Bosma, Fei Xia, Ed~Chi, Quoc~V Le, Denny Zhou, et~al. 2022{\natexlab{b}}.
\newblock Chain-of-thought prompting elicits reasoning in large language models.
\newblock \emph{Advances in Neural Information Processing Systems}.

\bibitem[{Yang et~al.(2023)Yang, Jin, Tang, Han, Feng, Jiang, Yin, and Hu}]{yang2023harnessing}
Jingfeng Yang, Hongye Jin, Ruixiang Tang, Xiaotian Han, Qizhang Feng, Haoming Jiang, Bing Yin, and Xia Hu. 2023.
\newblock Harnessing the power of llms in practice: A survey on chatgpt and beyond.
\newblock \emph{arXiv preprint arXiv:2304.13712}.

\bibitem[{Yu et~al.(2023)Yu, Jiang, Shi, Yu, Liu, Zhang, Kwok, Li, Weller, and Liu}]{metamath}
Longhui Yu, Weisen Jiang, Han Shi, Jincheng Yu, Zhengying Liu, Yu~Zhang, James~T Kwok, Zhenguo Li, Adrian Weller, and Weiyang Liu. 2023.
\newblock Metamath: Bootstrap your own mathematical questions for large language models.
\newblock \emph{arXiv preprint arXiv:2309.12284}.

\bibitem[{Yuan et~al.(2023)Yuan, Yuan, Li, Dong, Tan, and Zhou}]{gsm8k-rft}
Zheng Yuan, Hongyi Yuan, Chengpeng Li, Guanting Dong, Chuanqi Tan, and Chang Zhou. 2023.
\newblock Scaling relationship on learning mathematical reasoning with large language models.
\newblock \emph{arXiv preprint arXiv:2308.01825}.

\bibitem[{Yue et~al.(2023)Yue, Qu, Zhang, Fu, Huang, Sun, Su, and Chen}]{mammoth}
Xiang Yue, Xingwei Qu, Ge~Zhang, Yao Fu, Wenhao Huang, Huan Sun, Yu~Su, and Wenhu Chen. 2023.
\newblock Mammoth: Building math generalist models through hybrid instruction tuning.
\newblock \emph{arXiv preprint arXiv:2309.05653}.

\bibitem[{Zhang et~al.(2019)Zhang, Wang, Zhang, Dai, and Shen}]{zhang2019gap}
Dongxiang Zhang, Lei Wang, Luming Zhang, Bing~Tian Dai, and Heng~Tao Shen. 2019.
\newblock The gap of semantic parsing: A survey on automatic math word problem solvers.
\newblock \emph{IEEE transactions on pattern analysis and machine intelligence}.

\bibitem[{Zhao et~al.(2023)Zhao, Zhou, Li, Tang, Wang, Hou, Min, Zhang, Zhang, Dong et~al.}]{survey_llm}
Wayne~Xin Zhao, Kun Zhou, Junyi Li, Tianyi Tang, Xiaolei Wang, Yupeng Hou, Yingqian Min, Beichen Zhang, Junjie Zhang, Zican Dong, et~al. 2023.
\newblock A survey of large language models.
\newblock \emph{arXiv preprint arXiv:2303.18223}.

\bibitem[{Zhou et~al.(2022)Zhou, Nova, Larochelle, Courville, Neyshabur, and Sedghi}]{zhou2022teaching}
Hattie Zhou, Azade Nova, Hugo Larochelle, Aaron Courville, Behnam Neyshabur, and Hanie Sedghi. 2022.
\newblock Teaching algorithmic reasoning via in-context learning.
\newblock \emph{arXiv preprint arXiv:2211.09066}.

\end{thebibliography}
\bibliographystyle{acl_natbib}

\clearpage
\appendix

\section{Related Work and Background}
The field of Large Language Models (LLMs) has witnessed substantial advancements, yet the integration of mathematical reasoning within these models remains a challenging frontier. Existing researches in LLMs primarily focus on the natural language understanding and generation~\cite{wei2022emergent, yang2023harnessing}, with limited exploration in mathematical problem-solving. The complexity of mathematical problems, which requires not only numerical computation but also logical inference and the understanding of abstract concepts, still remains a notable challenge for LLMs~\cite{survey_llm, survey_math, survey_mlp, qian2022limitations, zhou2022teaching, verify, drori2021neural, zhang2019gap}. To address this challenge, many researches are being conducted via the following approaches: 1) prompting approach, 2) fine-tuning approach, and 3) continued pretraining approach.

\paragraph{Prompting Approach}
Recent studies are based on the prompting methods for mathematical reasoning without additional training. Recently, the concepts of Chain of Thoughts (CoT)~\cite{cot} and Program of Thoughts (PoT)~\cite{pot, pal} have emerged as promising approaches to enhance mathematical reasoning in LLMs.
The CoT involves breaking down complex reasoning problems into a series of intermediate reasoning steps. This approach has shown promise in improving the accuracy and reliability of LLMs in mathematical problem-solving, by mimicking the human thought process of step-by-step reasoning. However, it is not ideal for solving complex mathematical problems~\cite{pot}. To address this issue, the PoT introduces a more algorithmic perspective. Specifically, it expresses the reasoning steps as code and delegate computation steps to an code interpreter. This approach allows the LLMs to effectively deal with problems that require a combination of mathematical operations and logical reasoning, by structuring the problem-solving process in a programmatic manner.

\paragraph{Fine-tuning Approach}
More recently, many works~\cite{wizardmath, mammoth, metamath, tora} focus on the fine-tuning LLMs for mathematical reasoning tasks. WizardMath~\cite{wizardmath} proposed Reinforcement Learning from Evol-Instruct Feedback (RLEIF), which integrates supervised fine-tuning (SFT) and proximal policy optimization (PPO) for mathematical reasoning. MAmmoTH~\cite{mammoth} introduces a new hybrid instruction-tuning dataset called MathInstruct\footnote{\url{https://huggingface.co/datasets/TIGER-Lab/MathInstruct}}, which consists of CoT rationale and PoT rationale. MetaMath~\cite{metamath} proposed a new instruction-tuning dataset named MetaMathQA\footnote{\url{https://huggingface.co/datasets/meta-math/MetaMathQA}}, which is augmented by question bootstrapping methods. ToRA~\cite{tora} suggested a series of tool-integrated reasoning agents, which is fine-tuned on the tool-use trajectories (PoT rationale) datasets generated by prompting GPT-4.

\paragraph{Continued Pretraining Approach}
Some researches~\cite{minerva, llemma} continually pretrain a base model to specialize in the mathematical reasoning. Minerva~\cite{minerva} is a large language model pretrained on general natural language data and further trained on the scientific and mathematical data. Llemma~\cite{llemma} was also obtained through continued pretraining Code Llama~\cite{codellama} on their own collected data named Proof-Pile-2\footnote{\url{https://huggingface.co/datasets/EleutherAI/proof-pile-2}}.

In this paper, we focus on the fine-tuning approach by integrating the CoT and PoT learning. Motivated by \citet{dong2023abilities} that showed that the abilities of LLMs can be improved depending on the SFT strategy, we analyze how much performance can be improved depending on the SFT strategy from the perspective of solving challenging mathematical problems.

\section{Detailed Descriptions of Seed Datasets}
\label{appendix:a}

The detailed description of each seed dataset is as follows: 
\begin{longenum}
\item \textbf{GSM8K}~\cite{gsm8k}: It focuses on elementary-level math problems to evaluate abilities that handle logical reasoning and parse and interpret math questions presented in natural language;
\item \textbf{MATH}~\cite{math}: It includes a wide range of math problems, ranging from elementary arithmetic to advanced topics such as algebra, calculus, and geometry, which are challenging more than GSM8K;
\item \textbf{MetaMathQA}~\cite{metamath}: It is a dataset augmented through rephrasing question, forward-backward reasoning~\cite{fobar}, self-verification, and answer augmentation based on GSM8K and MATH;
\item \textbf{MathInstruct}~\cite{mammoth}: It consists of a mix of 13 types of CoT and PoT mathematical rationales from various mathematical fields. Specifically, CoT type data consist of GSM8K, GSM8K-RFT~\cite{gsm8k-rft}, AQuA-RAT~\cite{aqua-rat}, MATH, THeoremQA~\cite{theoremqa} Camel-Math~\cite{camel-math} and College-Math. Otherwise, PoT type data consist of GSM8K, AQuA-RAT, TheoremQA, MathQA~\cite{mathqa} and NumGLUE~\cite{numglue};
\item \textbf{QANDA}: It consists of a diverse collection of real-world mathematical questions and detailed solutions, catering to a broad spectrum of mathematical concepts and difficulty levels.
\end{longenum}

\end{document}